\documentclass[english]{bmvc2k}
\usepackage[latin9]{inputenc}
\usepackage{units}
\usepackage[numbers]{natbib}

\makeatletter

\title{\vspace*{-12mm}\begin{center}{Anisotropic Agglomerative Adaptive Mean-Shift}\end{center}}

\addauthor{Rahul Sawhney}{rahul.sawhney@gatech.edu}{1}
\addauthor{Henrik I. Christensen}{hic@gatech.edu}{1}
\addauthor{Gary R. Bradski}{gbradski@magicleap.com}{2}

\addinstitution{
 Institute of Robotics and Intelligent Machines\\
 Georgia Institute of Technology\\
 Atlanta, USA
}
\addinstitution{
 Magic Leap, Inc.\\
 Dania Beach, USA
}

\runninghead{S{awhney}, C{hristensen}, B{radski}}{A{nisotropic}  A{daptive} M{ean} S{hift}}

\usepackage{lipsum}
\usepackage[small,compact]{titlesec}

\usepackage{color}
\usepackage{colortbl}\usepackage{xcolor}

\usepackage{graphicx}
\usepackage{caption}
\usepackage{subcaption}

\usepackage[export]{adjustbox}
\usepackage[english]{babel}

\usepackage{setspace}
\usepackage{multirow}
\usepackage{tabularx}
\usepackage{booktabs}
\usepackage{comment}

\usepackage{algorithm}
\usepackage{algpseudocode}

\usepackage{relsize}

\usepackage{mdwlist}
\usepackage{enumitem}

\makeatletter
\newcommand*{\mycaptionfont}{\@setfontsize\mycaptionfont{8pt}{.85em}}
\makeatother

\DeclareCaptionFont{mycaptionfont}{\mycaptionfont}

\definecolor{figcapblue}{rgb}{0,0,0.4}
\DeclareMathSizes{10}{8}{7}{5}

\expandafter\def\expandafter\normalsize\expandafter{%
    \normalsize
    \setlength\abovedisplayskip{-2pt}
    \setlength\belowdisplayskip{3pt}
    \setlength\abovedisplayshortskip{-2pt}
    \setlength\belowdisplayshortskip{3pt}
}

\makeatother

\usepackage{babel}
\begin{document}
\maketitle

\vspace*{-2mm}
\begin{abstract}
Mean Shift today, is widely used for mode detection and clustering.
The technique though, is challenged in practice due to assumptions
of isotropicity and homoscedasticity. We present an adaptive Mean
Shift methodology that allows for full anisotropic clustering, through
unsupervised local bandwidth selection. The bandwidth matrices evolve
naturally, adapting locally through agglomeration, and in turn guiding
further agglomeration. The online methodology is practical and effecive
for low-dimensional feature spaces, preserving better detail and clustering
salience. Additionally, conventional Mean Shift either critically
depends on a per instance choice of bandwidth, or relies on offline
methods which are inflexible and/or again data instance specific.
The presented approach, due to its adaptive design, also alleviates
this issue - with a default form performing generally well. The methodology
though, allows for effective tuning of results.
\end{abstract}

\section{\label{sec:intro}Introduction}

\textquoteleft Mean Shift' (\citep{fukunaga1975estimation,cheng1995mean},
MS) is a powerful nonparametric technique for unsupervised pattern
clustering and mode seeking. References \citep{comaniciu2002mean,bradski1998real}
established it's utility in low-level perception tasks such as feature
clustering, filtering and in tracking. It has been in popular use
since, as a very useful tool for pattern clustering of sensor data
(\citep{shotton2013real,erdogmus2008information} for example). It
has also found niche as a preprocessor (a priori segmentation, smoothing)
before higher level image \& video analysis tasks such as scene parsing,
object recognition, detection (\citep{kohli2009robust,yang2007multiple,ke2007event}).
Image segmentation approaches such as Markov Random Fields, Spectral
clustering, Hierarchical clustering use it as an a priori segmenter
with improved results (\citep{kohli2009robust,ozertem2008mean,Kim2010LearningAffinities,unnikrishnan2007toward,vsurkala2011hierarchical}).

\begin{figure}[t!]
\begin{minipage}[t]{1\columnwidth}
\caption*{\label{fig:Motive:AAAMS}(\emph{a}) 3D Clustering result ($23$ clusters) over image data (left, \emph{L*a*b*} space) by the proposed approach. \emph{1-sigma} final trajectory bandwidths have been overlaid over the converged modes. The segment image is shown on right.}
\includegraphics[width=1\columnwidth]{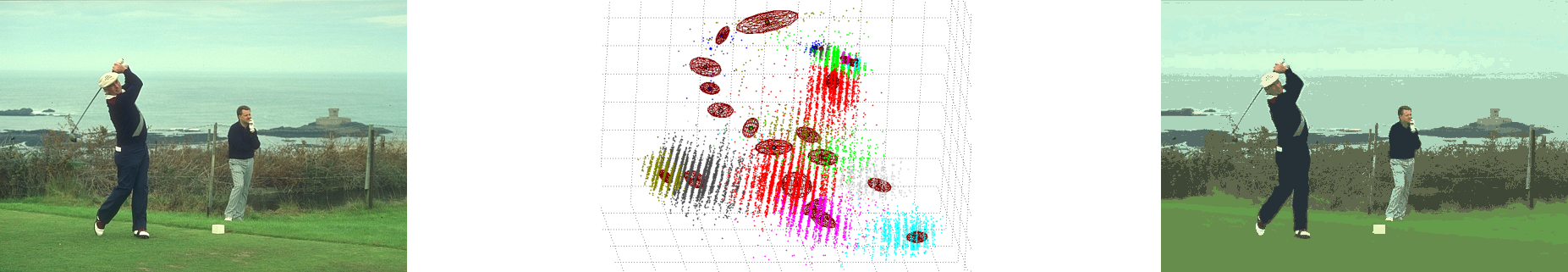}
\end{minipage}%
\vspace{2mm}
\begin{minipage}[b]{1\columnwidth}
\caption*{\label{fig:Motive:ConMS}(\emph{b}) Comparitive results with standard MS (left) and variable-bandwidth isotropic MS (\cite{bgeorgescu2003mean}, right), at similar clustering levels, $25$ {\scriptsize{}\&}{\scriptsize} $27$ respectively, are shown. Final mode locations have been indicated over the cluster plots. MS with correctly chosen bandwidth detected more coherent modes than \cite{bgeorgescu2003mean}, but looses partition saliency (bushes, water, sky in background). \cite{bgeorgescu2003mean} better adapts to scales but oversegments at places, and smooths over others (face). Both smoothed over details, failed to detect some modes at lower scales (trouser edges, maroon on shirt {\scriptsize{}\&}{\scriptsize} shoes). In general, conventional MS had a typical tendency to over-segment heavily or compromise partition boundaries.}
\includegraphics[width=1\columnwidth]{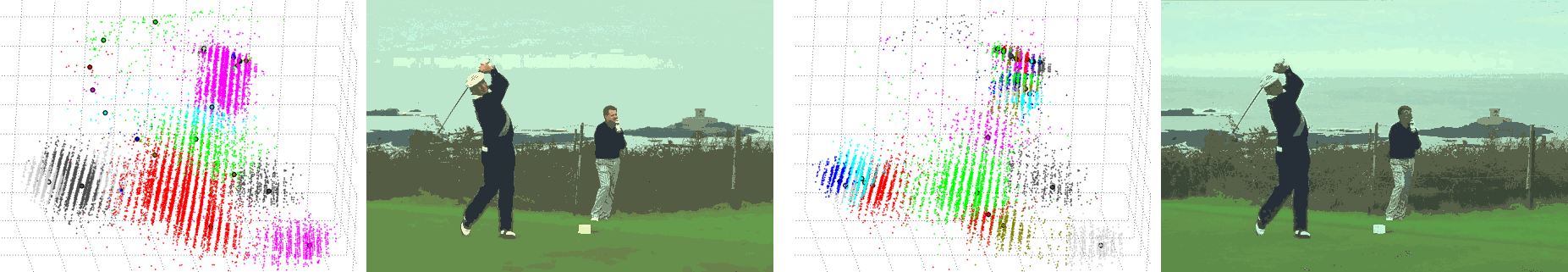}
\end{minipage}%
\caption{\label{fig:Motive} Exemplar illustrative result of our approach, AAAMS (a), is shown along with conventional MS results (b), at comparable clustering levels. As is indicated by the plots and segment images,  AAAMS effectively adapts to local scale and preserves anisotropic details, affecting more salient partitions.}
\end{figure}

Mean Shift methodologies though, employ some assumptions and have
some limitations, which may not be desirable. Its popular standard
form, \citep{comaniciu2002mean}, utilizes fixed, scalar bandwidth
assuming homoscedasticity and isotropicity. Being homoscedastic, it
also requires proper bandwidth choice on a per instance basis. The
adaptive Mean Shift variants, \citep{comaniciu2001variable,bgeorgescu2003mean},
ascertain variable bandwidths, but they still assume isotropicity.
They also make use of heuristics which are not flexible, and lack
clustering control. Offline bandwidth selection methods for Mean Shift
(\citep{chacon2013data,horova2013full,comaniciu2003algorithm}), typically
estimate a single, global bandwidth, and/or are data specific/non-automatic.
As\emph{ }indicated in\emph{ Fig.}\emph{\ref{fig:Motive}} - isotropic/scalar
bandwidths tend to smooth anisotropic patterns and affect partition
boundaries, while global/homoscedastic bandwidths are inappropriate
when clusters (or modes) at different scales need to be identified.

We present a Mean Shift methodology which is anisotropic and locally
adaptive. It is able to leverage guided agglomeration for unsupervised
bandwidth selection (\emph{Fig.}\emph{\ref{fig:Motive}}, {\color{red}\emph{5}}).
This results in robust mode detection, with increased partition saliency.
Also as a consequence, a low valued parameter set performs nicely
over a wider range of data instances (\emph{Sec.} \ref{sub:Update-Equations}).

Clusters arise on the fly in the proposed approach, as a consequence
of agglomeration of extant clusters. \emph{Local bandwidths }(\emph{Secs.
\ref{sub:MotivationBackground}}, \ref{sec:Methodology}) which evolve
anisotropically every iteration, are associated with each cluster;
by design, all members of a cluster converge to the same local mode.
By evolving as function of a cluster's aggregated trajectory points,
these bandwidths are able to adapt to the underlying mode structure
(shape, scale, orientation) - and in turn, guide future cluster trajectory
and agglomeration. The supplementary also presents a useful result
- a convergence proof when full bandwidths vary between Mean Shift
iterations, as is the case here. We refer to our approach as online
because it's an on the fly unsupervised procedure; with simple bookkeeping
doing away with re-calculations.

\subsection{\label{sub:MotivationBackground}Motivation and Background}

We utilize the exposition style of \citep{carreira2007gaussian}.
Let $\left\{ x_{i}\right\} _{i=1}^{n}\subset R^{d}$, be a set of
\emph{d}-dimensional data points with their sample point kernel density
estimate (KDE) being $p(x)=\sum_{i=1}^{n}p(x_{i})p(x|x_{i})=\sum_{i=1}^{n}p_{i}\frac{1}{c_{i}}K(\|x-x_{i}\|_{\Sigma_{i}})$.
Stationary points of the KDE can be estimated by evaluating the density
gradient and setting it to zero. This gives rise to the Mean-Shift
fixed point iteration :

\vspace*{-3mm}
\begin{subequations}
\begin{gather} 
x^{\text{\ensuremath{\tau}+1}}=f(x^{\tau}) \label{eq:fpt1} \\
f(x^{\tau})=\bigg(\sum_{i=1}^{n}p_{i}\frac{1}{c_{i}}K^{'}(\|x^{\tau}-x_{i}\|_{\Sigma_{i}})\Sigma_{i}^{-1}\bigg)^{-1}\times\bigg(\sum_{i=1}^{n}p_{i}\frac{1}{c_{i}}K^{'}(\|x^{\tau}-x_{i}\|_{\Sigma_{i}})\Sigma_{i}^{-1}x_{i}\bigg) \label{eq:fpt2} 
\end{gather}
\end{subequations}

\noindent$K(t),\, t\geq0$, is a \emph{d}-variate kernel with compact
support satisfying some regularity constraints, mild in practice (\citep{comaniciu2002mean,carreira2007gaussian}
for details). $\|x-x_{i}\|_{\Sigma_{i}}\equiv((x-x_{i})^{T}\Sigma_{i}^{-1}(x-x_{i}))^{\nicefrac{1}{2}}$,
is the Mahanalobis metric. The point prior $p_{i}\equiv p(x_{i})$~is
usually taken as $\nicefrac{1}{n}$. $c_{i}$ is a normalizing constant
depending only on the covariance matrix, $\Sigma_{i}$ (kernel bandwidth),
associated with each data point. The bandwidth, $\Sigma_{i}$, is
roughly an inverse measure of local curvature around $x_{i}$. It
linearly captures the scale and correlations of the underlying data.
$\tau$ indicates the iteration count. In practice, since $K(t)$
is taken with truncated support, the summations are only over $n'$
neighbors of $x^{\tau}$, with $n'\ll n$. The vector $m(x^{\tau})=f(x^{\tau})-x^{\tau}$,
is referred to as the Mean Shift. It\textquoteright s a bandwidth
scaled version of $\nabla p(x)$, is free from a step size parameter,
is large in regions with low $p(x)$ and small near the modes. Starting
at a data point, $x_{i}^{\tau=0}\equiv x_{i}$, the fixed point update
is run multiple times till convergence. The resulting points,~$x_{i}^{\tau\geq1}$
is referred to as the \emph{trajectory} of $x_{i}$, tracing a path
to the local mode. The technique thus, is able to locate modes and
partition feature space, without a priori knowledge of partition count
or structure.

The above hinges on selecting reasonable bandwidth matrices $\Sigma_{i}$.
Good bandwidths capture the underlying local distribution effectively.
In our approach, data points (pertaining to a cluster) converging
to a common local mode share a common bandwidth - one which reflects
this mode's structure, and to an extent, its basin of attraction (\citep{comaniciu2002mean}).
We refer to it as the local bandwidth (\citep{comaniciu2003algorithm}
utilizes local bandwidths in a related sense).

In online unsupervised usage, almost all Mean Shift variants for clustering,
for example \citep{comaniciu2002mean,vsurkala2011hierarchical,zhang2006accelerated,paris2007topological},
work under the restrictive assumptions of homoscedasticity and isotropicity
($\Sigma_{i}=\sigma^{2}I$, standard fixed bandwidth Mean Shift).
The scale parameter $\sigma$ has to be set carefully based on the
dataset instance. \citep{yuan2012agglomerative} utilizes set covering
based iterative agglomeration for improved efficiency. Coverage is
ensured through overlaps of small fixed homoscedastic bandwidths.
Some applications only assume isotropicity ($\Sigma_{i}=\sigma_{i}^{2}I$,
adaptive / variable-bandwidth Mean Shift). $\sigma_{i}$ is estimated
using a variation of the following two heuristics (\citep{comaniciu2001variable,bgeorgescu2003mean})
- 1) $k^{th}$ nearest neighbor, $x_{i}^{k}$, distance heuristic
$\rightarrow$~$\sigma_{i}\propto\| x_{i}-x_{i}^{k}\|$, or 2) Abramson\textquoteright s
heuristic $\rightarrow$ $\sigma_{i}\propto\sigma_{o}(\pi(x_{i}))^{\nicefrac{-1}{2}}$,
where $\pi(x)$ is the \emph{pilot} density estimate obtained by first
running mean shift with analysis bandwidth, $\sigma_{o}$. They have
found more use in smoothing type applications as reported in~\citep{mayer2009adaptive,Alaniz2006MRIAdaptiveMS}.
Variants have also been used in tracking scenarios, where the bandwidths
are adapted in a task specific fashion (see \citep{Jeong2005AdaptiveTracking,Comaniciu2003KernelTracking},~for
example). \citep{leibe2004scale,vojir2013robust} adapt isotropic
bandwidths to object scales, to unimodally track, search for them.
The topological, blurring, evolving variants for clustering (like
\citep{paris2007topological,vsurkala2011hierarchical,zhang2006accelerated,Carreira-Perpinan:2006:FNC:1143844.1143864,Surkala2012Evolving})
use isotropic bandwidths. They are primarily aimed at increased efficiency,
with results on par with standard mean shift. \citep{vedaldi2008quick}~presents
improvements over the somewhat related Mediod Shift. They propose
usage of their algorithm as initialization for Mean Shift, for increased
efficiency.

In offline settings,~\citep{comaniciu2003algorithm}~presents a
supervised methodology. Training data is processed with analysis bandwidths
to select local bandwidths based on neighboring partition stability.
The estimated bandwidths are then used to partition similarly distributed
test image data. Only recently were automatic full bandwidth selectors
for density gradient estimation proposed in~\citep{horova2013full,chacon2013data},
for offline settings. These focus on obtaining good data density gradients
(as opposed to clustering) and optimize based on the mean square integrated
error (MISE). A single global bandwidth is estimated for the given
data, and as the authors themselves note, the involved computations
are not straighforward.

A very useful variant is Joint Domain Mean Shift, \citep{comaniciu2002mean},
which is used to create partitions jointly respecting the dataset\textquoteright s
multiple feature domains which are mutually independent; For example,
$\left\langle color,space\right\rangle $ in color based segmentation
\& smoothing, and $\left\langle color,flow\right\rangle $ in motion
segmentation. When {\scriptsize{}$x=\left[{x^{r}}^{T}\, {x^{s}}^{T}\right]$}{\scriptsize}
with {\scriptsize{}$\left(x^{r}\,\bot\, x^{s}\right)|x$}{\scriptsize},
and utilizing two separate kernels, $K_{r},\, K_{s}$, we'll have
{\scriptsize{}$p(x)=\sum_{i=1}^{n}p(x_{i})p(x^{r}|x_{i})p(x^{s}|x_{i})$}{\scriptsize}.
\emph{Eq.}\ref{eq:fpt2} analogue would then come out to be {\scriptsize{}$f(x^{\tau})=\left(\sum_{i=1}^{n}p_{i}\frac{1}{c_{i}^{'}}J(\|x^{\tau,r}-{x_{i}}^{r}\|_{\Sigma_{i}^{r}},\|x^{\tau,s}-{x_{i}}^{s}\|_{\Sigma_{i}^{s}})\Sigma_{i}^{-1}\right)^{-1}\times\left(\sum_{i=1}^{n}p_{i}\frac{1}{c_{i}^{'}}J(\|x^{\tau,r}-{x_{i}}^{r}\|_{\Sigma_{i}^{r}},\|x^{\tau,s}-{x_{i}}^{s}\|_{\Sigma_{i}^{s}})\Sigma_{i}^{-1}x_{i}\right)$}{\scriptsize},
where $c_{i}^{'}$ is the normalization constant, {\scriptsize{}$J(t_{1},t_{2})\equiv K_{r}^{'}(t_{1})K_{s}(t_{2})=K_{r}(t_{1})K_{s}^{'}(t_{1}),\forall t_{1},t_{2}\ge0$}{\scriptsize},
and {\scriptsize{}$\Sigma_{i}=\left[\begin{array}{cc} \Sigma_{i}^{r} & 0\\ 0 & \Sigma_{i}^{s} \end{array}\right]$}{\scriptsize}.
Typically, but not necessarily,~$x^{s}$ may lie on a spatial manifold
- imposing structure to data which is utilized. Instances in literature
use fixed global scale parameters $\sigma^{r}$ and $\sigma^{s}$,
which have the aforementioned limitations.~As noted in \citep{unnikrishnan2007toward}
on color segmentation, $\sigma^{r}$ and $\sigma^{s}$ need to be
selected carefully. Good choices are not always possible, with segments
being too coarse or too fine at times (\emph{Figs. }\emph{\ref{fig:singleparamcomp}},
{\color{red}\emph{5}}). Reference \citep{wang2004image} utlizes
an anisotropic $\Sigma_{i}^{s}$~for visual data segmentations. Every
data point's associated bandwidth, $\Sigma_{i}$, is modulated multiple
times in each iteration, until convergence is achieved. Modulation
heuristics have been provided, to be deployed as per task. The spatial
bandwidth $\Sigma_{i}^{s}$~is parameterized as function of eigenvectors
of neighborhood data covariance.~$\Sigma_{i}^{r}$ is taken to be
an isotropic scalar dependent on~$\Sigma_{i}^{s}$.

\section{\label{sec:Methodology}Methodology}

A data point, $x_{i}$, is alternatively represented as $x_{i,u}$
- the first index value being its unique identifier as before and
the second index indicating its current, exclusive membership to a
cluster, $u\in\left\{ 1,\dots n\right\} $ %
\footnote{The second index is left out when the membership is apparent or inconsequential.
We similarly ease out the notations whenever pertinent, to simplify
exposition without loss of intuition.%
}. A cluster $u$'s constituent data points is denoted by the set ,
$C_{u}=\{x_{i,u}\mid\exists\, i\in\{1,\dots n\}\}$. By algorithm
design, clusters are merged only when they are tending towards the
same mode - thus all member points of a cluster, $u$, will eventually
converge to a common local mode, say~$\mu{}_{u}$. They hence, are
also taken to share a common local bandwidth,~$\Sigma{}_{u}$. This
bandwidth develops every iteration when the cluster $u$'s trajectory
points set, $T_{u}$ , gets additional elements. The set of clusters
surviving at iteration, $\tau$, would be $U^{\tau}=\left\{ u\mid C_{u}\neq\emptyset\right\} $.
$\left|U^{\tau}\right|$ would indicate its cardinality. At beginning,
at $\tau=0$, each point trivially forms a separate cluster $\rightarrow$
$U^{0}=\{1,\dots n\}$ , $C_{u}=\{x_{i=u,u}\},\,\forall u\in U^{0}$.
Given the initialization, each extant cluster $u\in U^{\tau}$ will
always contain the initial point, $x_{u,u}$ - which we refer to as
its \emph{principle member}. 

At any iteration $\tau$, for each extant cluster $u$, mean shift
updates happen for only the principle member,~$x_{u,u}$; with the
first iteration running over trivial clusters. The resulting trajectory
is specified as $x_{u,u}^{\tau\geq1}$ or simply $u^{\tau}$. A cluster's
trajectory might end when it gets merged or converged. In general,
each data point, $x_{i}$, started out as a trivial cluster, and had
or still has a trajectory - it's trajectory set being $\{x_{i}^{\tau=1:end}\}$.
$'end'$~being the iteration at which the trajectory ended; else
the current iteration. Note that the data point $x_{i}$ itself is
not included in this set. For any surviving cluster \emph{u}, then,
the complete set of all agglomerated trajectory points associated
with it, would be $T_{u}=\{\cup\,\{x_{i}^{\tau=1:end}\}\mid x_{i}\in C_{u}\}$
- basically a union of all the members' trajectory sets. $u^{\tau}$
is indicative of the cluster $u$'s location. At convergence, $u^{\tau}$
would be the location of a local mode. $u$'s members would then be
comprising of data points pertaining to that mode and its basin (\emph{Fig.~}{\color{red}\emph{1(a)}}).
The data density in the immediate vicinity of $u^{\tau}$'s current
position is indicated as $\rho(u^{\tau})$, or simply, $\rho_{u}$.
We use operator $\Pi$ to retrieve the cluster identifier of an arbitrary
data point; so $\Pi(x_{i,u})=u$. The $n'$ data points in $u^{\tau}$'s
neighborhood are denoted as $Ne_{x}(u^{\tau})$, and the clusters
containing them as $G=\left\{ \cup\,\Pi(y)\mid y\in Ne_{x}(u^{\tau})\right\} $.%
\footnote{Since a cluster corresponds one-to-one with its principle member,
principle member's trajectory is at times referred to as cluster trajectory.
Similarly, convergence of trajectory is at times referred to as cluster
converging. $\tau$, apart from indicating iteration, also differentiates
between a trajectory point and a data point. The cluster trajectory,
$u^{\tau}\equiv x_{u,u}^{\tau\geq1}$~, is the trajectory resulting
from data point $x_{u,u}$ . The cluster $u$'s current location refers
to current position of the principle member, indicated by $u^{\ensuremath{\tau}}$.%
}

\begin{figure}[t!]
\bgroup
\newcommand{\lwid}{.6pt}
\newcommand{\sepspaceone}{.5em}
\newcommand{\sepspacetwo}{.75em}
\newcommand{\sepspacethreewo}{.95em}
\emph{\scriptsize{}\textbf{Algorithm 1 : AAAMS - Anisotropic Agglomerative Adaptive Mean Shift}}
\\[-7pt]
\noindent\rule{129mm}{\lwid}\par
\begin{minipage}[t]{0.51\linewidth}
\fontsize{6.5pt}{10pt}\selectfont
\vspace{-1mm}
\noindent $Function\,:\, AAAMS\left\langle \left\{ x_{i}\right\} _{i=1}^{n}\right\rangle ~ with ~ x_{i}\in R^{d}$ \par
\noindent $Returns\,:\,\left\langle U^{*},\, C^{*},\,\{\mu_{u}\}_{u\in U^{*}},\,\{\Sigma_{u}^{*}\}_{u\in U^{*}}\right\rangle$ \par
\vspace{\sepspaceone}
\noindent {\tiny{}$\#\#~Convergence\, Criteria\,\rightarrow\,\left\Vert m_{u}\right\Vert \leq\delta$}{\tiny \par}
\noindent $U^{o}=\left\{1,\dots n\right\}~;~C_{u}=\left\{x_{u}\right\},~x_{u}^{o}=x_{u},~\Sigma_{u}=\sigma_{base}^{2}I_{d},~\forall u\in U^{0}$ \par
\noindent $\tau=0~;~\lambda=5~;~\delta=Convergence~epsilon$ \par
\noindent $m_{u}=Large\,\in\, R^{d},~T_{u}=\phi\,,\,\forall u\in U^{0}$ \par
\noindent ~~~~~~~~$While~~\exists u\,\in U^{\tau}s.t.\,\left\Vert m_{u}\right\Vert >\delta $~ \par
\noindent ~~~~~~~~~~~$ForEach~~u\,\in\, U^{\tau}\, s.t.\,\left\Vert m_{u}\right\Vert >\delta$ \par
\noindent ~~~~~~~~~~~~~~$u^{\tau+1}=
\begin{cases} \begin{array}{c} Eq.\,\ref{eq:ScalarUp}\\ Eq.\,\ref{eq:FullUp}\\  Eq.\,\ref{eq:PartialUp}\end{array} & \begin{array}{c} ESS(u)\,<\,\lambda\\ ESS(g)\,\geq\lambda,\,\forall g\in G\\ otherwise \end{array}\end{cases}$ \par
\noindent ~~~~~~~~~~~~~~$m_{u}=u^{\tau+1}-u^{\tau}\,;\, T_{u}=T_{u}\cup u^{\tau+1}$ \par
\noindent ~~~~~~~~~~~~~~$Get\, Ne_{x}(u^{\tau+1})$ \par
\noindent ~~~~~~~~~~~~~~$ForEach~~y\,\in\, Ne_{x}(u^{\tau+1})~or~till~C_{u}\neq\emptyset$ \par
\noindent ~~~~~~~~~~~~~~~~~$If~\Pi(y)\,=u~or~C_{\Pi(y)}=\phi\,\,\, Then\,\,\, Continue$ \par
\noindent ~~~~~~~~~~~~~~~~~$If~\left\Vert u^{\tau+1}-y\right\Vert >\epsilon\,\,\, Then\,\,\, Continue$ \par
\noindent ~~~~~~~~~~~~~~~~~$If~!MergeCheck\left\langle u^{\tau+1},y,m_{u},y^{\tau=1}-y,u,\Pi(y)\right\rangle$ \par
\noindent ~~~~~~~~~~~~~~~~~~~~$Then~Continue$ \par
\end{minipage}%
\hspace*{3mm}
\begin{minipage}[t]{0.45\linewidth}
\fontsize{6.5pt}{10pt}\selectfont
\vspace{-1mm}
\noindent ~~~~~~~~~~~~~~~~~$If~\rho_{u}>\rho_{\Pi(y)\,}$ \par
\noindent ~~~~~~~~~~~~~~~~~~~~$Then$ \par
\noindent ~~~~~~~~~~~~~~~~~~~~~~$C_{u}=C_{u}\cup C_{\Pi(y)}\,;\, C_{\Pi(y)}=\emptyset$ \par
\noindent ~~~~~~~~~~~~~~~~~~~~~~$T_{u}=T_{u}\cup T_{\Pi(y)}\,;\, T_{\Pi(y)}=\emptyset$ \par
\noindent ~~~~~~~~~~~~~~~~~~~~$Else$ \par
\noindent ~~~~~~~~~~~~~~~~~~~~~~$C_{\Pi(y)}=C_{u}\cup C_{\Pi(y)}\,;\, C_{u}=\emptyset$ \par
\noindent ~~~~~~~~~~~~~~~~~~~~~~$T_{\Pi(y)}=T_{u}\cup T_{\Pi(y)}\,;\, T_{u}=\emptyset$ \par
\noindent ~~~~~~~~~~~~~~~$EndForEach$ \par
\noindent ~~~~~~~~~~~~~~~$If~ C_{u}=\phi\,\,\,\, Then\,\,\, Continue$ \par
\noindent ~~~~~~~~~~~~~~~$\Sigma_{u}=\begin{cases} \begin{array}{c} Eq.\,\ref{eq:band}\\ \Sigma_{u} \end{array} & \begin{array}{c} ESS(u)\,\geq\,\lambda\\ otherwise \end{array}\end{cases}$ \par
\vspace{\sepspaceone}
\noindent {\tiny{}~~~~~~~~~~~~~~~\#\#~$Optionally~Perturb\left\langle u^{\tau+1},\, m_{u}\right\rangle~~if~\left\Vert m_{u}\right\Vert \leq\delta$}{\tiny \par}
\noindent ~~~~~~~~~~~$EndForEach$ \par
\noindent ~~~~~~~~~~~$U^{\tau+1}=\left\{ u\mid u\,\in\, U^{\tau},\, C_{u}\neq\emptyset\right\}$ \par
\noindent ~~~~~~~~~~~$\tau=\tau+1$ \par
\noindent ~~~~~~~~$EndWhile$ \par
\noindent ~~~~~~~$U^{*}=U^{\tau}$~;~~$C^{*}=C_{u}\,,\,\forall u\in U^{\tau}$~; ~~$\Sigma_{u}^{*}=\Sigma_{u}\,,\,\forall u\in U^{\tau}$ \par
\noindent $EndFunction$ \par
\end{minipage}
\\[4pt]
\hspace*{40mm}\rule{42mm}{\lwid}\\[2pt]
\noindent \emph{\scriptsize{}For feature spaces that can be decomposed into independent subspaces, the above can be extended to multiple domains. The update equations would then utlize multiple kernels. Basically, for each domain, a }{\tiny{}$\left\langle \sigma_{base},\,\epsilon\right\rangle$}\emph{\scriptsize{} pair needs to be set.}{\scriptsize \par}
\vspace{1mm}
\noindent \emph{\scriptsize{}For example, for joint domain Mean Shift (Sec:\emph{\ref{sub:MotivationBackground})}, we'll have }{\tiny{}$\left\langle \sigma_{base}^{r},\,\epsilon^{r}\right\rangle \,\&\,\left\langle \sigma_{base}^{s},\,\epsilon^{s}\right\rangle$}\emph{\scriptsize{} for the two domains.~We'll have then }{\tiny{}$\Sigma_{base}=\bigg[\begin{array}{cc} \sigma_{base}^{r^{2}}I_{r} & 0\\ 0 & \sigma_{base}^{s^{2}}I_{s} \end{array}\bigg]~\&~\Sigma_{u}=\left[\begin{array}{cc} \Sigma_{u}^{r} & 0\\ 0 & \Sigma_{u}^{s} \end{array}\right]$. }\emph{{\tiny{}$\Sigma_{u}^{r}~\&~\Sigma_{u}^{s}$}{\tiny}\scriptsize{}~would~be~evaluated~from Eq.\emph{\ref{eq:band}}.~Eq.\emph{\ref{eq:FullUp}}~analogue~would~be}{\tiny{}$~f(u^{\tau})=\bigg(\mathlarger{\sum}_{\forall g\in G}\frac{1}{c_{g}}\Sigma_{g}^{-1}\mathlarger{\sum}_{\forall i\mid x_{i,g}\in Ne_{x}(u^{\tau})}J(\|u^{\tau,r}-{x_{i}}^r\|_{\Sigma_{u}^{r}},~\|u^{\tau,s}-{x_{i}}^s\|_{\Sigma_{u}^{s}})\bigg)^{-1}\times\bigg(\mathlarger{\sum}_{\forall g\in G}\frac{1}{c_{g}}\Sigma_{g}^{-1}\mathlarger{\sum}_{\forall i\mid x_{i,g}\in Ne_{x}(u^{\tau})}J(\|u^{\tau,r}-{x_{i}}^r\|_{\Sigma_{u}^{r}},~\|u^{\tau,s}-{x_{i}}^s\|_{\Sigma_{u}^{s}})x_{i}\bigg)$;~}\emph{\scriptsize{}likewise for others.}{\scriptsize\par}
\noindent\rule{129mm}{\lwid}\par
\caption*{}
\egroup
\vspace*{-3mm}
\end{figure}

The methodology for anisotropic, agglomerative, adaptive Mean Shift
(AAAMS) is presented as a pseudo code in \emph{Alg}. {\color{red}\emph{1}}.
At every iteration, the following steps are run for each surviving
cluster that has not converged $\rightarrow$

\hspace*{-\parindent} 
\begin{minipage}[h!]{1\textwidth}
\vspace*{2mm}
\fontsize{7.5pt}{0pt}\selectfont
\begin{itemize}[leftmargin=*,itemsep=2pt,topsep=4pt,parsep=0pt,partopsep=0pt]
\item[1)] Mean shift update is computed and the cluster's location is updated. No merges happen before the first update.  
\item[2)] Nearest neighbors about the current location are ascertained - they are utilized for cluster merges, and for the mean shift update in subsequent iteration.  
\item[3)] When merge criteria are met, either some clusters (owners of the neighborhood points which lie within epsilon) get merged into this cluster, or this cluster gets merged into one of them.  
\item[4)] If the incumbent cluster survived after the merge, its bandwidth is updated.  
\item[5)] Optionally, if the cluster has converged, its location could be perturbed a bit. It is, then, not taken out of consideration in subsequent iteration.
\end{itemize}
\end{minipage}

\subsection{\label{sub:Update-Equations}Update Equations}

Taking $p_{i}=\nicefrac{1}{n}$~and limiting summations to the neighboring
points, $Ne_{x}(u^{\tau})$, the fixed point iteration, \emph{Eq.}
\emph{\ref{eq:fpt1}-\ref{eq:fpt2}}, over a cluster $u$ (rather $x_{u,u}$)
can be reformulated/reorganized as a local bandwidth based decomposition
:\begin{subequations}
\begin{gather} 
u^{\tau+1}=f(u^{\tau}),\, where\, u^{\tau=0}\,\equiv\, x_{u,u} \label{eq:fpt_reform} \\
f(u^{\tau})=\bigg(\sum_{\forall g\in G}\frac{1}{c_{g}}\Sigma_{g}^{-1}\sum_{\forall\iftrue i\mid\fi x_{i,g}\in Ne_{x}(u^{\tau})}K^{'}(\|u^{\tau}-x_{i}\|_{\Sigma_{g}})\bigg)^{-1}\times\bigg(\sum_{\forall g\in G}\frac{1}{c_{g}}\Sigma_{g}^{-1}\sum_{\forall\iftrue i\mid\fi x_{i,g}\in Ne_{x}(u^{\tau})}K^{'}(\|u^{\tau}-x_{i}\|_{\Sigma_{g}})\, x_{i}\bigg) \label{eq:FullUp}
\end{gather}
\end{subequations}

\emph{Eq.} \emph{\ref{eq:FullUp}} would be exactly the same as \emph{Eq.}
\emph{\ref{eq:fpt2}} at $\tau=0$, when all points form trivial clusters.
When local homoscedasticity in neighborhood of $u^{\tau}$ is assumed
with the cluster's own bandwidth $\Sigma_{u}$ taken as bandwidth
estimate for neighborhood $Ne_{x}(u^{\tau})$, \emph{Eq.} \emph{\ref{eq:FullUp}}
simplifies %
\footnote{\emph{Eq.}\ref{eq:PartialUp} gets us a particularly insightful interpretation.
Note that $\|u^{\tau}-x_{i}\|_{\Sigma_{u}}$could be thought of as
a partial likelihood measure of the data point $x_{i}$~belonging
to the cluster~$u$. Consider the conditional $\rightarrow\, p(\nicefrac{x_{i}}{u^{\tau}};u)=\nicefrac{K^{'}(\|u^{\tau}-x_{i}\|_{\Sigma_{u}})}{\sum_{\dots}K^{'}(\|u^{\tau}-x_{i}\|_{\Sigma_{u}})}$
, with the summation in denominator normalizing the distribution.
The fixed point update\emph{ }from\emph{ Eq.}\ref{eq:PartialUp} would
then come out to be $u^{\tau+1}=\sum_{\dots}p(\nicefrac{x_{i}}{u^{\tau}};u)x_{i}$.
So the updated cluster trajectory $u^{\tau+1}$~is just the neighborhood
data expectation, conditioned only under the cluster's own distribution.
In effect, this serves to guide/update a cluster's trajectory based
only on the properties (bandwidth) it has itself ascertained (till
$\tau$).%
} to :

\begin{equation}
f(u^{\tau})=\frac{\sum_{\forall i\mid x_{i}\in Ne_{x}(u^{\tau})}K^{'}(\|u^{\tau}-x_{i}\|_{\Sigma_{u}})x_{i}}{\sum_{\forall i\mid x_{i}\in Ne_{x}(u^{\tau})}K^{'}(\|u^{\tau}-x_{i}\|_{\Sigma_{u}})}\label{eq:PartialUp}
\end{equation}

If global homoscedasticity and isotropicity is assumed, \emph{Eq.}
\emph{\ref{eq:FullUp}} takes the form of standard mean shift update,
where the bandwidth is specified through a fixed scalar $\sigma_{base}$
:

\begin{equation}
f(u^{\tau})=\frac{\sum_{\forall i\mid x_{i}\in Ne_{x}(u^{\tau})}K^{'}(\|\nicefrac{\left(u^{\tau}-x_{i}\right)^{T}\left(u^{\tau}-x_{i}\right)}{\sigma_{base}^{2}}\|)\, x_{i}}{\sum_{\forall i\mid x_{i}\in Ne_{x}(u^{\tau})}K^{'}(\|\nicefrac{\left(u^{\tau}-x_{i}\right)^{T}\left(u^{\tau}-x_{i}\right)}{\sigma_{base}^{2}}\|)}\label{eq:ScalarUp}
\end{equation}

Each trivial cluster utilizes fixed base bandwidth to begin with,
employing \emph{Eq.} \emph{\ref{eq:ScalarUp}} for mean shift updates.
Benign clusters form and start moving up on some modes. As soon as
a cluster accumulates enough trajectory points for full bandwidth
estimates (\emph{Sec.}\ref{sub:Bandwidth-Estimation}) to be significant
($u$ has moved up to denser regions by then), it switches to anisotropic
updates, given by \emph{Eqs.} \emph{\ref{eq:PartialUp}}~\&~\emph{\ref{eq:FullUp}}.
A reasonable test of significance for $\Sigma_{u}$ estimates, is
to check if the kernel weighted point count or \emph{Effective Sample
Size} ($ESS$, \citep{duong2008feature}) is above some value, $\lambda$.

\begin{equation}
ESS(u)=\frac{\sum_{\forall v\in T_{u}}K^{'}(\left\Vert \overline{T_{u}}-v\right\Vert _{\Sigma_{u}^{estimate}})}{\sum_{\forall v\in T_{u}}K^{'}(\left\Vert 0\right\Vert _{\Sigma_{u}^{estimate}})}\label{eq:esslabel}
\end{equation}

$\overline{T_{u}}$ indicates the mean of the trajectory set. The
anisotropic update \emph{Eq.} \emph{\ref{eq:PartialUp}} is used when
the cluster has an $ESS(u)\geq\lambda$ , and the more confident update
\emph{Eq.} \emph{\ref{eq:FullUp}} is used, when $ESS(g)\geq\lambda,\,\forall g\in G$
- when all the neighboring clusters too have confident enough bandwidth
estimates %
\footnote{We note empirically for dense data, as in images, a simple cluster
size sufficiency check works well. For joint domains, a cluster could
switch to anisotropic updates when it has atleast $\max\left(dim(x^{r}),\, dim(x^{s})\right)^{2}$
members.%
}. As a binomial rule of thumb (\citep{duong2008feature}), $\lambda=5$
is chosen as the minimum $ESS$, which is analogous to choosing 5
as the minimum individual expected cell counts in a $\chi^{2}$ test
of independence.

So starting with the initial base scalar, $\sigma_{base}$, the bandwidth
matrices evolve by themselves. The nice part is that just a low base
value suffices for reasonably dense data, with the bandwidths scaling
data driven thereon and adapting to the local structure's scale, shape
and orientation. $\sigma_{base}$~thus becomes indicative of the
minimum desired detail in the data space. This is opposed to traditional
Mean Shift - where the bandwidth scalar is indicative of the scale
at which the data space has to be partitioned.

\subsection{\label{sub:Bandwidth-Estimation}Bandwidth Estimation}

Bandwidth estimates based on a cluster's member data point locations
are not reliable (\citep{comaniciu2003algorithm} notes this too).
A subset of point locations in isolation cannot be considered as representative
of underlying distribution. The underlying local distribution is actually
a localized subset of the joint non-parametric density represented
by the entire dataset - it has significant contributions from neighboring
structures as well. The local structure could also be asymmetric and/or
without tail(s). A solution lies in considering points which arise
from mean shift ascents over the mode the cluster is converging to
- the cluster trajectory set, $T_{u}$. We use the variance of $T_{u}$
with respect to the underlying density as an estimate, $\Sigma_{u}$.
As $T_{u}$ builds up each iteration, so does $\Sigma_{u}$.

\begin{equation}
\Sigma_{u}=\frac{\sum_{\forall v\in T_{u}}\rho(v)vv^{\tau}}{\sum_{\forall v\in T_{u}}\rho(v)}-\eta_{u}\eta_{u}^{T}+\,\xi I,\,\, where\,\eta_{u}=\frac{\sum_{\forall v\in T_{u}}\rho(v)v}{\sum_{\forall v\in T_{u}}\rho(v)}\label{eq:band}
\end{equation}

$\rho(v)$~is the data density in the immediate vicinity of a point
$v\in T_{u}$. This is evaluated using $\sigma_{base}$ for consistency
across clusters. $\eta_{u}\,\&\,\Sigma_{u}$~are then basically the
expectation and variance of the localized distribution. In practice,
a small regularizer,~$\xi$, has to be added to the diagonals of
$\Sigma_{u}$~to prevent degenerate fitting in sparse regions, and
for numerical stability.\iftrue While computing anisotropic updates,
eigenvalue decomposition is employed and any eigenvalues of $\Sigma_{u}$
which fall below~$\xi$~, are clamped to it.\fi~Note that $\Sigma_{u}$
always remains positive definite. Also note that all summations are
computed on the fly. 

\emph{Eq.} \ref{eq:band}~could also be thought of as density weighted
trajectory set variance. As a cluster approaches a mode, mean shift
trajectory points get more concentrated and are weighted more, leading
to a conservative but more localized and robust estimate \textendash{}
more immune to long tails.\textbf{ }\emph{Figs. \emph{\ref{fig:Motive}}, \color{red}{\emph{5}}
}plot the bandwidths and modes at convergence, for color and point
data.

\begin{figure}[t!]
\parindent0pt
\newcommand{\sepspacebelow}{-.5em}
\newcommand{\sepspaceabove}{.25em}
\centering
\begin{subfigure}[h]{1\linewidth}
{\caption{\label{fig:variation:qual} {Effects of varying the detail and vicinity parameters on a brush painting with smudged colors.\hfill} \vspace*{\sepspacebelow}}
\includegraphics[width=1\columnwidth]{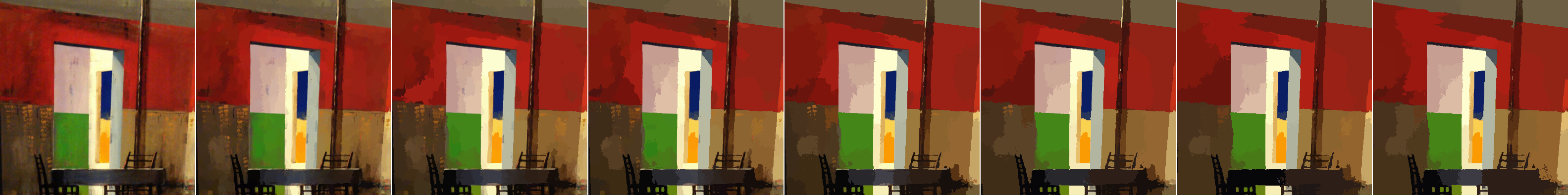}\\[-2em]
\noindent{\flushleft{}\tiny{}\emph{Acrylic \& Oil}
~~~~~~~~~~~~~~~~~~~~~$\bigl\langle4,9,.5,16\bigr\rangle$
~~~~~~~~~~~~~~~~$\bigl\langle9,16,.75,25\bigr\rangle$
~~~~~~~~~~~~~~$\bigl\langle16,25,1,36\bigr\rangle$
~~~~~~~~~~~~~~$\bigl\langle25,36,2,81\bigr\rangle$
~~~~~~~~~~~~~~$\bigl\langle36,49,1.5,64\bigr\rangle$
~~~~~~~~~~$\bigl\langle49,64,1.5,100\bigr\rangle$
~~~~~~~~~~$\bigl\langle64,81,1,121\bigr\rangle$~~~~~~~~}{\tiny\flushleft}\hfill%
}\end{subfigure}\hfill
\vspace*{-1.5em}
\begin{subfigure}[h]{1\linewidth}
{\caption{\label{fig:variation:quant} {Parameter sensitivity plots. Each of {\tiny{}$\bigl\langle{\sigma_{base}^{r}}^{2},{\sigma_{base}^{s}}^{2},\,\epsilon_{r}^{2},\,\epsilon_{s}^{2}\bigr\rangle$}{\tiny} was varied while keeping others constant. Their effects on number of clusters, their average size,  and iterations for convergence are plotted.  Results were averaged over{\scriptsize{} $33$}{\scriptsize} images. As with conventional MS, color domain parameters are understandably more sensitive.{\scriptsize{} $\delta = .01$}{\scriptsize} was used.} \vspace*{-.3em} }
\includegraphics[width=1\columnwidth]{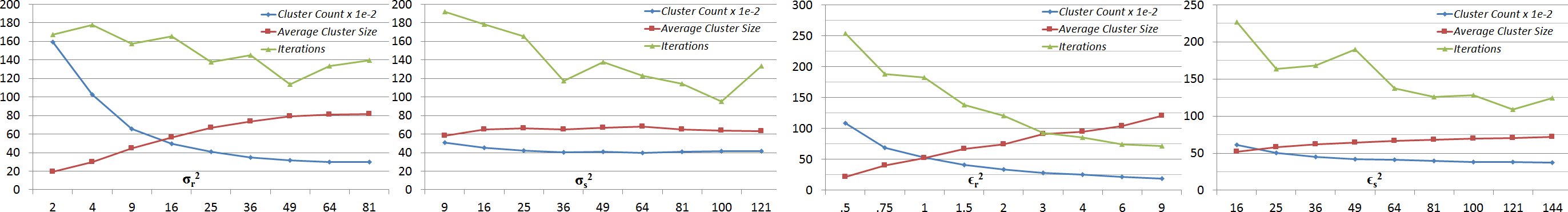}
}\end{subfigure}\hfill
\vspace*{\sepspaceabove}
\protect\caption{\label{fig:variation} {For joint domain AAAMS over images, we show the qualitative and quantitative effects of varying the detail and vicinity parameters, {\tiny{}$\bigl\langle{\sigma_{base}^{r}}^{2},{\sigma_{base}^{s}}^{2},\,\epsilon_{r}^{2},\,\epsilon_{s}^{2}\bigr\rangle$}{\tiny}. Post processing was disabled, except for enforcing cluster contiguity. As can be seen, if the need be, a good control over smoothing and segmentation levels can be exercised.}}
\end{figure}%

\subsection{\label{sub:Cluster-Merging}Cluster Merging}

\vspace{-1mm}

For any given data points, if their mean shift trajectories intersect,
they will converge to a common local mode. Thus in the vicinity of
a data point's trajectory (which is moving up some mode) - any data
points in sufficient proximity, having their shift vectors deemed
to be intersecting with this trajectory, could be clustered together.
They will eventually end up converging on the same local mode. So
we basically consider the data points in the vicinity of a cluster
trajectory, $u^{\tau}$ - with an epsilon~$\epsilon$, delineating
the vicinity. If a data point, $y$, in vicinity is ascertained (in
\emph{MergeCheck}) to be heading to the same mode as $u^{\tau}$,
then by transitivity - all the members of its parent cluster, $\Pi(y)$,
are heading to that mode too - the clusters $u$ and $\Pi(y)$ , can
then be merged. The cluster which is higher up the mode (higher density)
assimilates the other cluster into itself, thus accelerating convergence
to the mode. This also helps in avoiding spurious merges. 

\textbf{\emph{MergeCheck}} - This is intentionally specified as a
generic function returning a true/false value. It could be implemented
to suit different feature spaces and clustering criteria. The more
holistic this check is, the larger the operating range of $\epsilon$~can
be (assuming the distance norm holds up), without impacting clustering
stability. In our experiments, we used a very lightweight generic
implementation that worked well over considered data spaces - basically
verifying through inner product checks that 1)~relative distance
between $u^{\tau+1}$ and $y$ is decreasing and 2)~Mean shift bearings
\footnote{The bearing at $u^{\tau+1}$ is $m_{u}$. The bearing at $y$, given
by~$y^{\tau=1}-y$ , is the mean shift vector resulting from the
first iteration over the trivial cluster containing $y$; it's stored
up for consequent use.%
} at $u^{\tau+1}$ and $y$ are in the same direction.~We note though
that divergence measures like Bhattacharya (\emph{Sec}. \ref{sub:Post-Processing}),
kernel induced feature space metrics (\citep{ozertem2008mean}), information-theoretic
ones like Renyi's entropy (\citep{erdogmus2008information}) seem
viable, interesting possibilities for MergeCheck. We are yet to experiment
with them.

\subsection{\label{sub:Post-Processing}Post Processing}

Once data has been partitioned, a post processing step merges clusters
with proximate modes, and ensures a minimum cluster size (in conventional
Mean Shift, clusters are delineated only during the post process).
Additionally for structured data, cluster contiguity could be enforced.
We use graph operations. For structured data as in images, adjacency
connections between clusters can be added naturally using a spatial
grid structure. For unstructured data, connections between a cluster
and all clusters within a reasonably large distance threshold (mode
to mode distances) were added, to ensure a connected graph. \emph{Bhattacharya}
divergence (\citep{bhattacharyya1946measure}, $d_{B}$) was used
as the merging criteria. It takes into account not just the variance
normalized mode proximity, but also the disparity in variances themselves
(\emph{Mahanalobis} measure is its special case). $0\leq d_{B}\leq4$
was a good range, with $d_{B}=1$ (somewhat analogous to $1-sigma^{2}$
disparity) performing well generally %
\footnote{For images, since color similarity alone is of consequence, $d_{B}$
was evaluated only over the $L^{*}a^{*}b^{*}$ space%
}. \emph{Alg}. {\color{red}\emph{2}} specifies the steps.

\begin{figure}[t!]
\bgroup
\begin{minipage}[t!]{0.49\linewidth}
\scriptsize
{\flushleft{} \emph{\textbf{Algorithm 2 : Post Processing}}} {\flushleft}
\vspace*{-4mm}\noindent\rule{62mm}{0.6pt}
\begin{itemize}[leftmargin=*,topsep=2pt,itemsep=0pt,parsep=2pt]
\item (For structured data only) For each cluster, use spatial adjacency to ascertain the disconnected components (highest density/mode locations for these small disconnected point sets need to be recomputed). Each disconnected component forms an additional separate cluster thereon.\hfill
\item Build the adjacency graph.\hfill
\item Merge all clusters which fall below minimum desired size, to the closest adjacent cluster until no such remain.\hfill
\item For each remaining cluster, using its constituent points, compute the density weighted variances, similar to \emph{Eq}.\ref{eq:band} - this is representative of the cluster's stand-alone distribution and alleviates tail influences.
\item For each pair of remaining clusters $\{a,\, b\}$, connected by an adjacency edge, evaluate ~$\rightarrow$ {\tiny{} $d_{B}=\frac{1}{8}\left(\mu_{a}-\mu_{b}\right){}^{T}\left(\frac{\Sigma_{a}+\Sigma_{b}}{2}\right)^{-1}\left(\mu_{a}-\mu_{b}\right)+\frac{1}{2}ln\left(\frac{det\left(\frac{\Sigma_{a}+\Sigma_{b}}{2}\right)}{\sqrt{det\left(\Sigma_{a}\right).det\left(\Sigma_{b}\right)}}\right)\label{bhatta}$}{\scriptsize{}. If it falls below a certain threshold, merge the two}.
\end{itemize}
\end{minipage}%
\hspace*{2mm}
\begin{minipage}[t!]{0.49\linewidth}
\tiny
\begin{tabular*}{1\textwidth}{@{\extracolsep{\fill}}|p{1.7cm}|p{3.47mm}|p{3.47mm}|p{3.47mm}|p{3.47mm}|}
\hline \textbf{\emph{Methods / Score}} & \textbf{\emph{PRI}} & \textbf{\emph{GCE}} & \textbf{\emph{VoI}} & \textbf{\emph{BDE}}
\tabularnewline \hline  $AAAMS$ & ${\color{red}.8230}$ & ${\color{red}.1589}$ & 2.1785 & ${\color{green}12.60}$
\tabularnewline \hline  $JMS^{*}$ & .7870 & ${\color{green}.1608}$ & 2.2484 & 13.34
\tabularnewline \hline  \textbf{\emph{Prior Art \cite{Kim2010LearningAffinities}}} & \textbf{\emph{}} & \textbf{\emph{}} & \textbf{\emph{}} & \textbf{\emph{}}
\tabularnewline \hline  $FullSpectralOverMS$~\cite{Kim2010LearningAffinities} & ${\color{green}0.8146}$ & 0.1809 & ${\color{green}1.8545}$ & ${\color{red}12.21}$
\tabularnewline \hline  $JMS$~\cite{christoudias2002synergism} & ${\color{blue}0.7958}$ & 0.1888 & ${\color{blue}1.9725}$ & 14.41
\tabularnewline \hline  $NCut$~~\cite{Kim2010LearningAffinities}\emph{~-~}Ref.\emph{~}{[}27{]} & 0.7330 & 0.2662 & 2.6137 & 17.19
\tabularnewline \hline  $MNCUT$~~\cite{Kim2010LearningAffinities}\emph{~-~}Ref.~{[}6{]} & 0.7632 & 0.2234 & 2.2789 & ${\color{blue}13.17}$
\tabularnewline \hline  $GBIS$~~\cite{Kim2010LearningAffinities}\emph{~-~}Ref.~{[}9{]} & 0.7139 & ${\color{blue}0.1746}$ & 3.3949 & 16.67
\tabularnewline \hline  $Saliency$~~\cite{Kim2010LearningAffinities}\emph{~-~}Ref.~{[}8{]} & 0.7758 & 0.1768 & ${\color{red}1.8165}$ & 16.24
\tabularnewline \hline  $JSEG$~~\cite{Kim2010LearningAffinities}\emph{~-~}Ref.~{[}7{]} & 0.7756 & 0.1989 & 2.3217 & 14.40
\tabularnewline \hline
\end{tabular*}
\caption*{\label{tab:bsdtable}{\scriptsize{}Table 1: Results on BSD300 \cite{martin2001database}. We used a single parameter set {\tiny{}$\bigl\langle20,36,1,64\bigr\rangle$}{\tiny} for AAAMS. For better results, {\tiny{}$d_{B}$}{\tiny} was set from {\tiny{}$\left\{.25,.5,1,1.25,1.5,2\right\}$}{\tiny}. JMS$^{*}$ parameters were selected per image to maintain similar segmentation levels, with an eye on preserving details, segment saliency.\\[0pt]
For perspective, we also reproduce results from \cite{Kim2010LearningAffinities} of unsupervised image segmentation methods. \cite{Kim2010LearningAffinities} selects segment levels per image. Top three values for each index are colored as ${\color{red}r}{\color{green}g}{\color{blue}b}$. AAAMS performs best overall - it's clearly ahead in PRI \& GCE, and is a close second in BDE. Note that \cite{Kim2010LearningAffinities}, which has the next best values, operates over \emph{a priori} Mean Shift segmentations.} {\scriptsize} }
\end{minipage}
\caption*{}
\egroup
\vspace*{-1mm}
\end{figure}

\section{Results}

\vspace{-1mm}

The base scalar parameter $\sigma_{base}$, in effect, regulates the
minimum desired detail in the feature space, the smoothing level.
The vicinity parameter,~$\epsilon$, regulates cluster merge chances
and hence cluster sizes. For images, with AAAMS operating over joint
domains of $\left\langle color,\, space\right\rangle $, the detail
and vicinity parameters would be {\scriptsize{}$\left\langle \sigma_{base}^{r},\,\sigma_{base}^{s}\right\rangle$}{\scriptsize}
and {\scriptsize{}$\left\langle \epsilon_{r},\,\epsilon_{s}\right\rangle$}{\scriptsize}
respectively (indicated in \emph{Alg}. {\color{red}\emph{1}}).\emph{
Fig}.~\emph{\ref{fig:variation}}, shows quantitative and qualitative
effects of their variation. Although a good degree of control is possible
to achieve a desired result, our experiments showed that any low valued
set gave nice results over a good range of images. 

Due to agglomeration, the number of clusters decrease monotonically
every iteration. Only a fraction of clusters remain after the first
couple of iterations; with the cluster count falling rapidly in all
early iterations. The scheme thus results in a drastic reduction in
net mean shift computes - as compared to the hitherto style of clustering
only after convergence, where computations happen for every data point,
in each iteration. (for dense image data, typically less than $5\%$
of the clusters remain by the $11^{th}$ or $12^{th}$ iteration).
This serves to offset the additional computational workload arising
from the use of full bandwidth matrices. Our straight up joint domain
implementation was achieving similar timings on average to standard
Mean Shift, which uses scalar bandwidths. Improvements in efficency
based on fast nearest neighbor search such as exploiting grid structure
of spatial domain, locally sensitive hashing (\citep{bgeorgescu2003mean})
are applicable in our methodology too. Using Gaussian kernels, with
a convergence delta,~$\delta$, set adequately to $.01$, merges
would cease before $90^{th}$ iteration, with convergence around the
$100^{th}$. When just pre-partitioning is the end objective, the
merging scheme thus allows us to fine tune stopping criteria. Along
with the first iteration shift vectors, globally normalized local
density values at each data point were stored for consequent use too.
In each iteration,~$\rho(u^{\tau})$ was then approximated by the
density value at $u^{\tau}$'s nearest data point. We found perturbations
to be generally useful, lending to mode detection robustness and more
salient partitioning. A cluster at convergence can be perturbed a
fixed number of times consecutively, with progressively damped magnitudes.~$u$
then, would not be brought out of contention in the next iteration
- although the immediate trajectory point resulting from the perturbation
will not be included in $T_{u}$. The results presented in this paper
though, are with perturbations disabled.

\begin{figure}[t!]
\begin{minipage}[t!]{1\columnwidth}
\scriptsize
\vspace*{-.5mm}
\noindent{~~~~~~~~~~~~~\emph{Image}~~~~~~~~~~~~~~~~~~~~~~~~~~~~~~~~~~\emph{JMS}~~~~~~~~~~~~~~~~~~~~~~~~~~~~~~~~~\emph{AAAMS}~~~~~~~~~~~~~~~~~~~~~~~~~~~~\emph{JMS Labels}~~~~~~~~~~~~~~~~~~~\emph{AAAMS Labels}\hfill}\\[-1pt]%
\noindent\includegraphics[width=1\columnwidth]{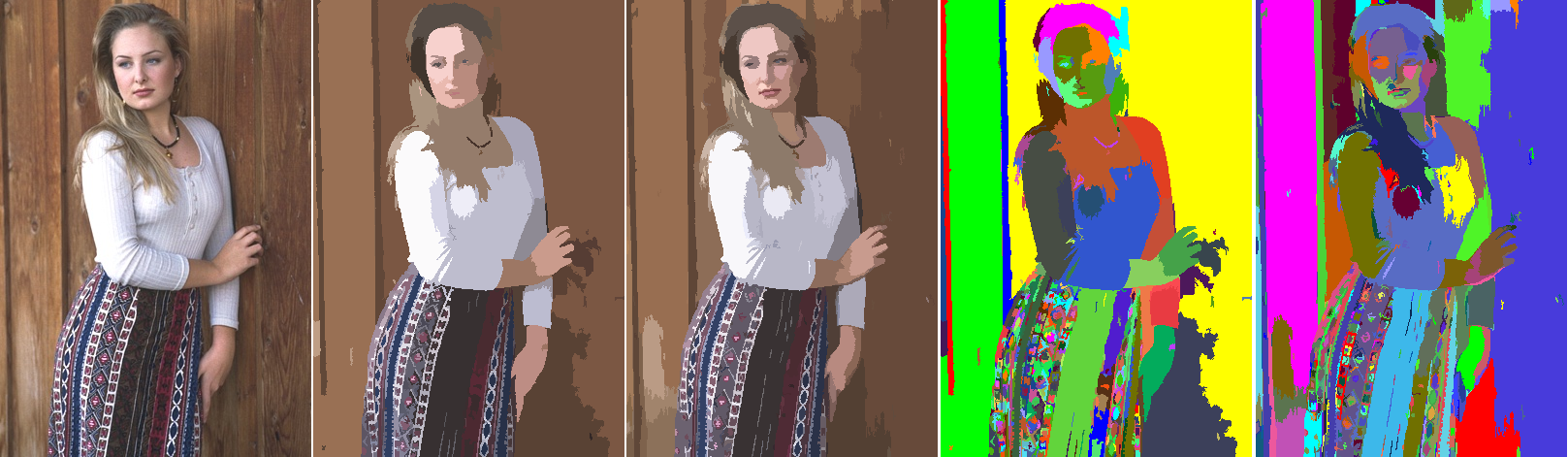}\\[-1pt]
\noindent\raggedright{}\emph{\hspace*{10mm}Lady~~~~~~~~~~~~~~~~~~~~~~JMS - 523 Labels,~~~~~~~~~~~~~~~~AAAMS - 490 Labels}
\noindent\includegraphics[width=1\columnwidth]{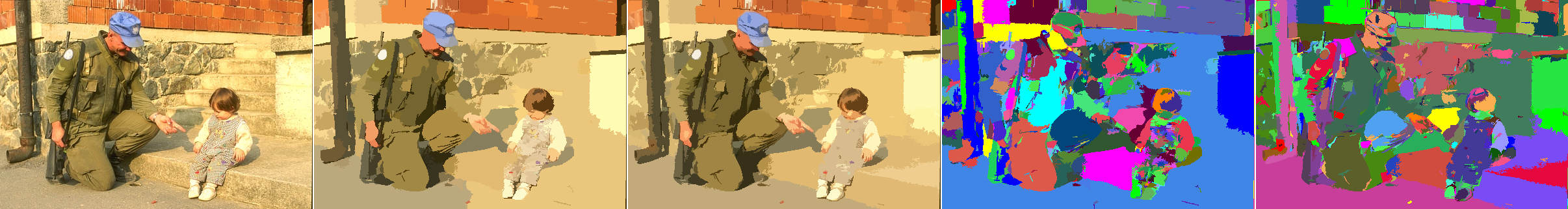}\\[-1pt]
\noindent\raggedright{}\emph{\hspace*{10mm}Soldier~~~~~~~~~~~~~~~~~~~~~~JMS - 617 Labels,~~~~~~~~~~~~AAAMS - 601 Labels}
\noindent\includegraphics[width=1\columnwidth]{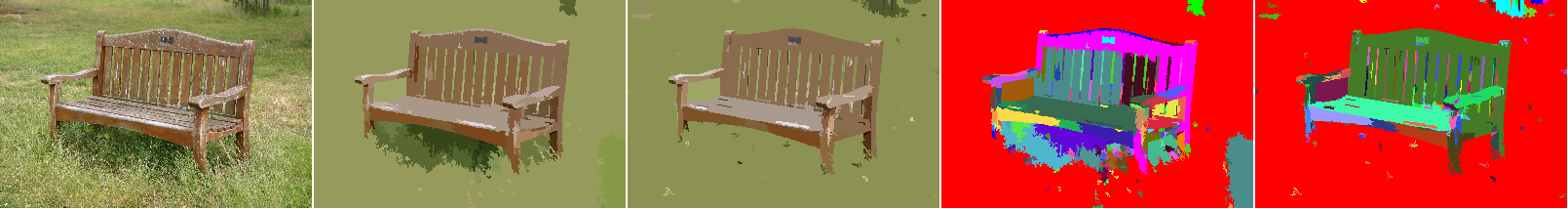}\\[-1pt]
\noindent\raggedright{}\emph{\hspace*{10mm}Bench~~~~~~~~~~~~~~~~~~~~~~~JMS - 180 Labels,~~~~~~~~~~~~AAAMS - 165 Labels}
\noindent\includegraphics[width=1\columnwidth]{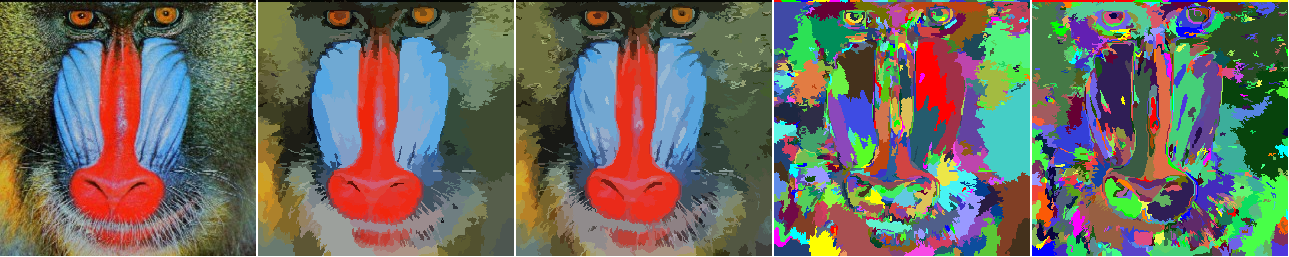}\\[-1pt]
\noindent\raggedright{}\emph{\hspace*{10mm}Mandrill~~~~~~~~~~~~~~~~~~JMS - 752 Labels,~~~~~~~~~~~~~AAAMS - 721 Labels}
\caption{\label{fig:singleparamcomp}AAAMS preserves more details and affects more perceptually salient segmentations, at similar clustering levels. We used a single parameter set,~{\tiny{}$\bigl\langle{\sigma_{base}^{r}}^{2},{\sigma_{base}^{s}}^{2},\,\epsilon_{r}^{2},\,\epsilon_{s}^{2}\bigr\rangle=\bigl\langle15,16,1,81\bigr\rangle$}{\tiny} with {\tiny{}$d_{B}=1$}{\tiny}, to show its adaptivity on varied images. JMS segments were kept around the same, with eye on preserving detail; it still smooths over at places. Its parameter values varied significantly from image to image - {\tiny{}$\,{\sigma^{r}}^{2}\in\left[49,\,81\right]\,,\,{\sigma^{s}}^{2}\in\left[100,\,289\right]$}{\tiny}. Minimum cluster size was {\tiny{}10}{\tiny}.}
\end{minipage}
\end{figure}

\begin{figure}[h]
\begin{minipage}[t!]{0.49\columnwidth}
\vspace*{-.4mm}
\tiny
\noindent\begin{raggedright}~~~~~~~\emph{Image}~~~~~~~~~~~~~~~~~~~~~\emph{JMS}~~~~~~~~~~~~~~~~~~~~\emph{AAAMS}~~~~~~~~~~~~~~~~\emph{JMS~Labels}~~~~~~~~~\emph{AAAMS~Labels} 
\par\end{raggedright}\vspace*{0pt}
\noindent\includegraphics[width=1\columnwidth]{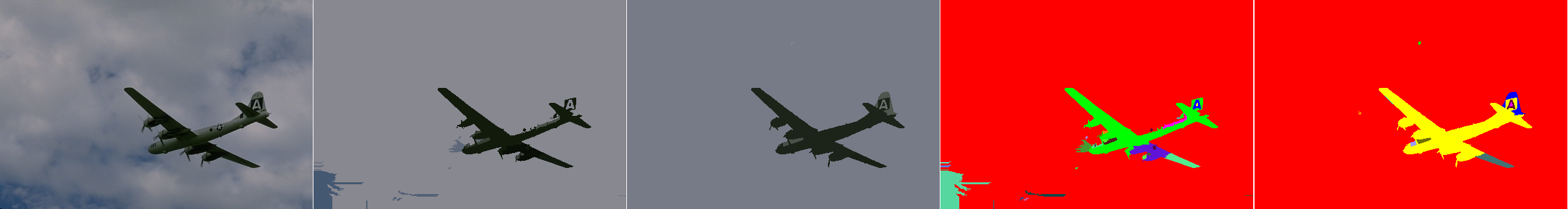}\\[-1pt]
\noindent\raggedright{}\emph{\hspace*{3mm}Aeroplane~~~~~~~~~~~~~~~JMS - 14 Labels,~~~~~~~~~~~~AAAMS - 11 Labels}
\noindent\includegraphics[width=1\columnwidth]{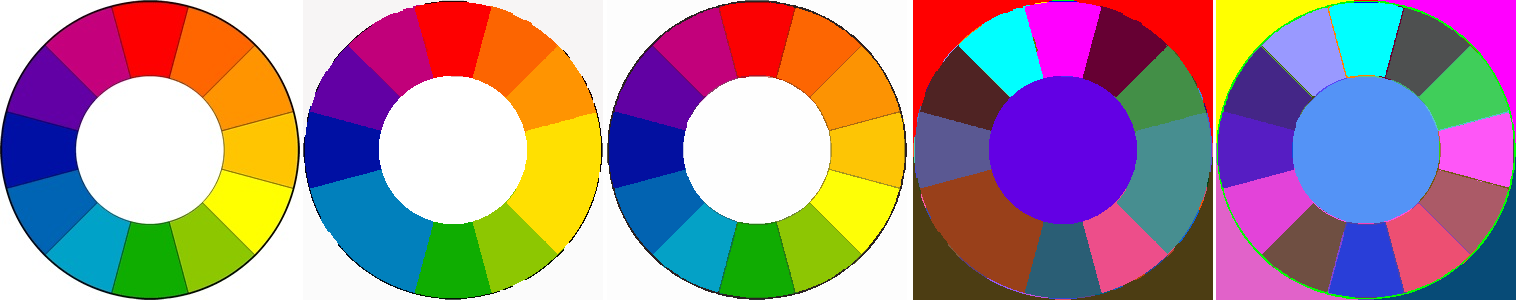}\\[-1pt]
\noindent\raggedright{}\emph{\hspace*{3mm}ColorWheel~~~~~~~~~~~~~~~JMS - 51 Labels,~~~~~~~~~~~~AAAMS - 48 Labels}
\noindent\includegraphics[width=1\columnwidth]{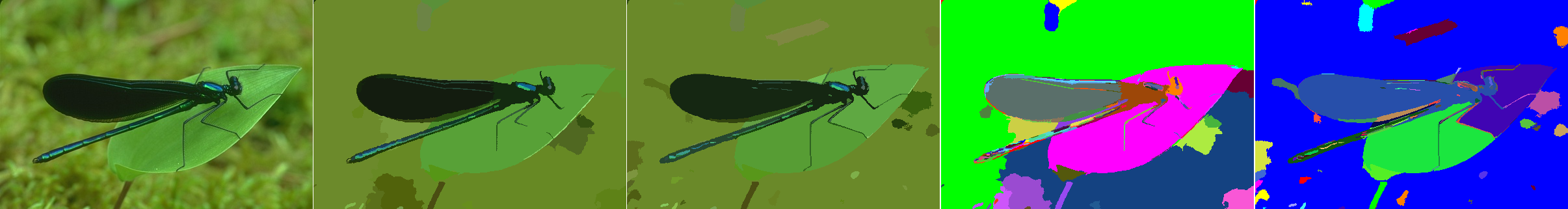}\\[-1pt]
\noindent\raggedright{}\emph{\hspace*{3mm}Wasp~~~~~~~~~~~~~~~JMS - 150 Labels,~~~~~~~~~~~~AAAMS - 111 Labels}
\caption*{\label{fig:parsimonycomp}\scriptsize{}Figure 4: More parsimonious segmentations were quite often not achievable with JMS - some varied examples are shown above (Images such as \emph{Lady} in \emph{Fig.3} are a typical case too). Both methods were configured for reduced label usage. Minimum cluster size was {\tiny{}10}{\tiny}. JMS, at its limit, is breaking boundaries and under segmenting. AAAMS with lesser labels, does not break boundaries, still maintains segment saliency.}
\end{minipage}
\hspace*{1mm}
\begin{minipage}[t!]{0.48\linewidth}
\noindent\includegraphics[width=1\columnwidth]{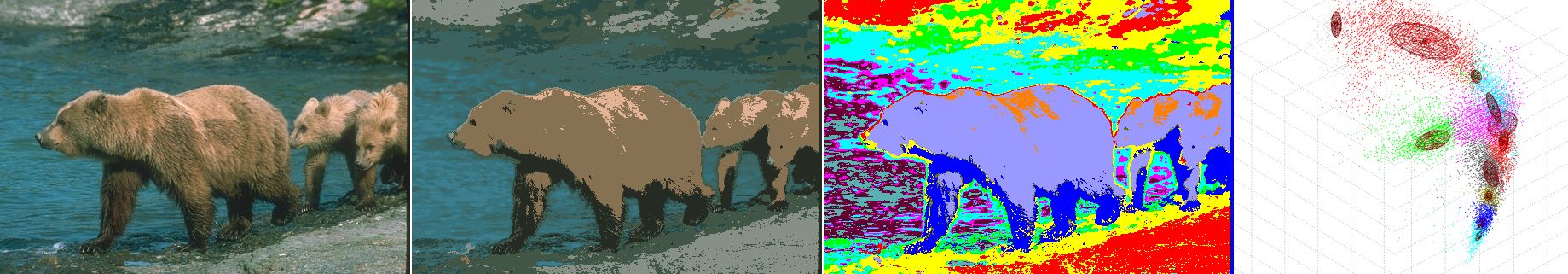}\\[2pt]
\noindent\includegraphics[width=1\columnwidth]{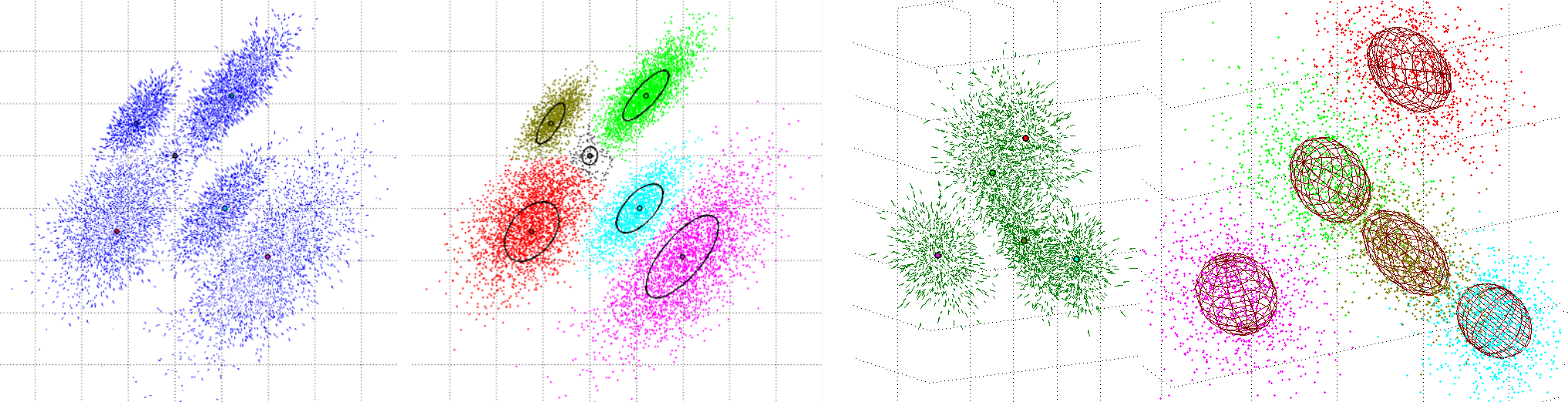}\\[-12pt]
\vspace*{-1mm}
\caption*{\label{fig:singledomain}\scriptsize{}Figure 5: Single domain clustering examples over color data (top row, {\tiny{}11}{\tiny} clusters) and  simulated gaussian mixtures (second row) in 2D  \& 3D respectively. $1-sigma$ final trajectory-set bandwidths have been overlaid at converged mode positions.\vspace*{1mm}}
\tiny
\begin{tabular*}{1\columnwidth}{@{\extracolsep{\fill}}p{17mm}|p{9mm}|p{9mm}|p{12mm}}
\hline  {$\mathbf{Data~\langle\#Dims,\#Classes\rangle}$}& \textbf{\emph{PRI}}& \textbf{\emph{GCE}}& \textbf{\emph{VoI}} 
\tabularnewline \hline  $Seeds~~~~~\langle7D,3\rangle$ & ${\color{red}.89}$ \textbf{/} .86 \textbf{/} .87 & ${\color{red}.17}$ \textbf{/} .20 \textbf{/} .19 & ${\color{red}0.85}$ \textbf{/} 0.98 \textbf{/} 0.93
\tabularnewline \hline  $Yeast~~~~~\langle8D,10\rangle$ & ${\color{red}.69}$ \textbf{/} .61 \textbf{/} .67 & .44 \textbf{/} ${\color{red}.39}$ \textbf{/} .47 & ${\color{red}3.03}$ \textbf{/} 3.10 \textbf{/} 3.22 
\tabularnewline \hline  $Letters~~~\langle16D,26\rangle$ & ${\color{red}.87}$ \textbf{/} .86 \textbf{/} .83 & .67 \textbf{/} .70 \textbf{/} ${\color{red}.62}$ & 4.96 \textbf{/} 5.16 \textbf{/} ${\color{red}4.72}$ 
\tabularnewline \hline  
\end{tabular*}
\caption*{\label{tab:bsdtable}{\scriptsize{}Table 2: Results on higher dimension real world datasets from \cite{asuncion2007uci}, with a single kernel. Indicated values are of AAAMS \textbf{/} MS \textbf{/} VariableMS (\cite{bgeorgescu2003mean}) respectively, with best values in ${\color{red}red}$.}{\scriptsize}}
\end{minipage}
\end{figure}

For image data, comparisons (\emph{Figs.}~\emph{\ref{fig:singleparamcomp}}, \emph{{\color{red}{4}}},
\emph{Table}.~\emph{{\color{red}{1}}} %
\footnote{Probabilistic Rand Index (PRI), Variation of Information (VoI), Global
Consistency Error (GCE), Boundary Displacement Error (BDE). The first
three are clustering purity measures. PRI is a measure of the fraction
of pairs of points whose labels are consistent with a given labeling.
VoI and BDE are relative distance metrics between two given segmentations,
based on average conditional entropy and boundary pixel difference,
respectively. GCE measures the extent to which one labeling can be
viewed as a refinement of the other. Higher is better for PRI while
lower is better for the other three. For BSD300, the values indicate
how well a segmentation corresponds to ones by human subjects. We
noticed that coarser segmentatios tended to give better values. This,
we suppose, was because humans tend to utilize much more comprehensive
cues, and incorporate object or more holistic level semantics in their
segmentations. It was noticed that PRI corresponded better to low
level segment saliency than others.%
}) are shown with joint domain Mean Shift implementation (JMS) from
EDISON (\citep{christoudias2002synergism}), over Berkely Segmentation
Dataset (\citep{martin2001database}, BSD300). BSD300 is meant for
supervised algorithms - we simply clubbed the training and test images
together. For sake of completeness, prior art on unsupervised image
segmentation is also shown in \emph{Table}.~\emph{{\color{red}{1}}}.
All indicated parameter values for AAAMS and JMS are squared. We did
not search for the best performing parameter set for AAAMS, opting
for a single low valued set instead. AAAMS performed significantly
better than JMS, with results superior to other unsupervised image
segmentation methods as well. 

Our experiments indicated that low base bandwidths, {\scriptsize{}$\left\langle \sigma_{base}^{r},\,\sigma_{base}^{s}\right\rangle$}{\scriptsize},
performed generally well on a good range of images (\emph{Fig}.~\emph{\ref{fig:singleparamcomp}}).
This was due to the presented approach being locally adaptive and
anisotropic. At similar clustering levels, AAAMS preserved more details
and affected more salient segmentations.

Single kernel AAAMS was tested on images and 2D, 3D gaussian mixtures
at varied scales - with nice results. AAAMS results in\emph{ Figs.}~\emph{\color{red}{1(a)}, {\color{red}{5}}}
are with postprocessing disabled. As\emph{ }indicated in \emph{Figs.}~\emph{\color{red}{1(a)}, {\color{red}{5}}},
reasonable local bandwidths arise, robustly identifying modes and
salient clusters, by adapting according to local structure.

Experiments were conducted with some higher dimension datasets from
\citep{asuncion2007uci} as well. \emph{Table.}~{\color{red}\emph{2}}
shows initial results, along with comparisons with single domain standard
Mean Shift (MS), and \citep{bgeorgescu2003mean}'s isotropic variable
bandwidth implementation. Cluster count was kept the same as class
count. AAAMS post-processing was disabled. \citep{bgeorgescu2003mean}
first determines isotropic point bandwidths using the $k^{th}$ nearest
neighbor distance heuristic, and subsequently utlizes them in single
kernel mean shift iterations. Our experiments with it indicated a
lack of clustering control. The datasets were meant for supervised
classification, with attributes/feature components at different scales,
and having uncorrelated and/or uninformative dimensions. Without any
pre-processing (normalizations, component analysis) decent results
were attained with a single kernel AAAMS. Note that \citep{bgeorgescu2003mean}
internally normalizes the data, while AAAMS \& MS results are without
any normalizations.

Promising results, both qualitative and quantitative, are indicative
of the efficacy of the presented approach. We intend to experiment
further, especially with different merging schemes and on varied data
spaces.

\section{Conclusion}

A generalized methodology for feature space partitioning and mode
seeking was presented - leveraging synergism of adaptive, anisotropic
Mean Shift and guided agglomeration. Unsupervised adaptation of full
anisotropic bandwidths is useful and further enables Mean Shift clustering.
We are excited about its prospects on point-normal clouds and video
streams.

Our experiments did indicate sparse data to be an issue. This is understandable,
as it encumbers cluster growth and bandwidth development, with AAAMS
behaving like conventional Mean Shift then. Future work would also
focus on alleviating this issue.



\bibliographystyle{plainnat}
\bibliography{egbib}

\begin{thebibliography}{37}
\providecommand{\natexlab}[1]{#1}
\providecommand{\url}[1]{\texttt{#1}}
\expandafter\ifx\csname urlstyle\endcsname\relax
  \providecommand{\doi}[1]{doi: #1}\else
  \providecommand{\doi}{doi: \begingroup \urlstyle{rm}\Url}\fi

\bibitem[Asuncion and Newman(2007)]{asuncion2007uci}
Arthur Asuncion and David Newman.
\newblock {UCI} machine learning repository - http://archive.ics.uci.edu/ml/,
  2007.

\bibitem[Bhattacharyya(1946)]{bhattacharyya1946measure}
Anil Bhattacharyya.
\newblock On a measure of divergence between two multinomial populations.
\newblock \emph{Sankhy{\=a}: The Indian Journal of Statistics}, pages 401--406,
  1946.

\bibitem[Bradski(1998)]{bradski1998real}
Gary~R Bradski.
\newblock Real time face and object tracking as a component of a perceptual
  user interface.
\newblock In \emph{Applications of Computer Vision, 1998. WACV'98.
  Proceedings., Fourth IEEE Workshop on}, pages 214--219. IEEE, 1998.

\bibitem[Carreira-Perpi\~{n}\'{a}n(2006)]{Carreira-Perpinan:2006:FNC:1143844.1143864}
Miguel~\'{A}. Carreira-Perpi\~{n}\'{a}n.
\newblock Fast nonparametric clustering with gaussian blurring mean-shift.
\newblock In \emph{Proceedings of the 23rd International Conference on Machine
  Learning}, ICML '06, pages 153--160, New York, NY, USA, 2006. ACM.

\bibitem[Carreira-Perpi\~{n}\'{a}n(2007)]{carreira2007gaussian}
Miguel~\'{A}. Carreira-Perpi\~{n}\'{a}n.
\newblock Gaussian {Mean-Shift} is an {EM} algorithm.
\newblock \emph{Pattern Analysis and Machine Intelligence, IEEE Transactions
  on}, 29\penalty0 (5):\penalty0 767--776, 2007.

\bibitem[Chac{\'o}n et~al.(2013)Chac{\'o}n, Duong, et~al.]{chacon2013data}
Jos{\'e}~E Chac{\'o}n, Tarn Duong, et~al.
\newblock Data-driven density derivative estimation, with applications to
  nonparametric clustering and bump hunting.
\newblock \emph{Electronic Journal of Statistics}, 7:\penalty0 499--532, 2013.

\bibitem[Cheng(1995)]{cheng1995mean}
Yizong Cheng.
\newblock Mean shift, mode seeking, and clustering.
\newblock \emph{Pattern Analysis and Machine Intelligence, IEEE Transactions
  on}, 17\penalty0 (8):\penalty0 790--799, 1995.

\bibitem[Christoudias et~al.(2002)Christoudias, Georgescu, and
  Meer]{christoudias2002synergism}
Christopher~M Christoudias, Bogdan Georgescu, and Peter Meer.
\newblock Synergism in low level vision.
\newblock In \emph{Pattern Recognition, 2002. Proceedings. 16th International
  Conference on}, volume~4, pages 150--155. IEEE, 2002.

\bibitem[Comaniciu et~al.(2003)Comaniciu, Ramesh, and
  Meer]{Comaniciu2003KernelTracking}
D.~Comaniciu, V.~Ramesh, and P.~Meer.
\newblock Kernel-based object tracking.
\newblock \emph{Pattern Analysis and Machine Intelligence, IEEE Transactions
  on}, 25\penalty0 (5):\penalty0 564--577, May 2003.

\bibitem[Comaniciu(2003)]{comaniciu2003algorithm}
Dorin Comaniciu.
\newblock An algorithm for data-driven bandwidth selection.
\newblock \emph{Pattern Analysis and Machine Intelligence, IEEE Transactions
  on}, 25\penalty0 (2):\penalty0 281--288, 2003.

\bibitem[Comaniciu and Meer(2002)]{comaniciu2002mean}
Dorin Comaniciu and Peter Meer.
\newblock Mean shift: A robust approach toward feature space analysis.
\newblock \emph{Pattern Analysis and Machine Intelligence, IEEE Transactions
  on}, 24\penalty0 (5):\penalty0 603--619, 2002.

\bibitem[Comaniciu et~al.(2001)Comaniciu, Ramesh, and
  Meer]{comaniciu2001variable}
Dorin Comaniciu, Visvanathan Ramesh, and Peter Meer.
\newblock The variable bandwidth mean shift and data-driven scale selection.
\newblock In \emph{Computer Vision, 2001. ICCV 2001. Proceedings. Eighth IEEE
  International Conference on}, volume~1, pages 438--445. IEEE, 2001.

\bibitem[Duong et~al.(2008)Duong, Cowling, Koch, and Wand]{duong2008feature}
Tarn Duong, Arianna Cowling, Inge Koch, and MP~Wand.
\newblock Feature significance for multivariate kernel density estimation.
\newblock \emph{Computational Statistics \& Data Analysis}, 52\penalty0
  (9):\penalty0 4225--4242, 2008.

\bibitem[Erdogmus et~al.(2008)Erdogmus, Ozertem, and
  Lan]{erdogmus2008information}
Deniz Erdogmus, Umut Ozertem, and Tian Lan.
\newblock Information theoretic feature selection and projection.
\newblock In \emph{Speech, Audio, Image and Biomedical Signal Processing using
  Neural Networks}, pages 1--22. Springer, 2008.

\bibitem[Fukunaga and Hostetler(1975)]{fukunaga1975estimation}
Keinosuke Fukunaga and Larry Hostetler.
\newblock The estimation of the gradient of a density function, with
  applications in pattern recognition.
\newblock \emph{Information Theory, IEEE Transactions on}, 21\penalty0
  (1):\penalty0 32--40, 1975.

\bibitem[Georgescu et~al.(2003)Georgescu, Shimshoni, and
  Meer]{bgeorgescu2003mean}
Bogdan Georgescu, Ilan Shimshoni, and Peter Meer.
\newblock Mean shift based clustering in high dimensions: A texture
  classification example.
\newblock In \emph{Computer Vision, 2003. Proceedings. Ninth IEEE International
  Conference on}, pages 456--463. IEEE, 2003.

\bibitem[Horov{\'a} et~al.(2013)Horov{\'a}, Kol{\'a}{\v{c}}ek, and
  Vopatov{\'a}]{horova2013full}
Ivana Horov{\'a}, Jan Kol{\'a}{\v{c}}ek, and Kamila Vopatov{\'a}.
\newblock Full bandwidth matrix selectors for gradient kernel density estimate.
\newblock \emph{Computational Statistics \& Data Analysis}, 57\penalty0
  (1):\penalty0 364--376, 2013.

\bibitem[Jeong et~al.(2005)Jeong, You, Oh, Oh, and
  Han]{Jeong2005AdaptiveTracking}
Mun-Ho Jeong, Bum-Jae You, Yonghwan Oh, Sang-Rok Oh, and Sang-Hwi Han.
\newblock Adaptive mean-shift tracking with novel color model.
\newblock In \emph{Mechatronics and Automation, 2005 IEEE International
  Conference}, volume~3, pages 1329--1333 Vol. 3, 2005.
\newblock \doi{10.1109/ICMA.2005.1626746}.

\bibitem[Jimenez-Alaniz et~al.(2006)Jimenez-Alaniz, Pohi-Alfaro,
  Medina-Bafluelos, and Yaflez-Suarez]{Alaniz2006MRIAdaptiveMS}
R.J. Jimenez-Alaniz, M.~Pohi-Alfaro, V.~Medina-Bafluelos, and O.~Yaflez-Suarez.
\newblock Segmenting brain mri using adaptive mean shift.
\newblock In \emph{Engineering in Medicine and Biology Society, 2006. EMBS '06.
  28th Annual International Conference of the IEEE}, pages 3114--3117, 2006.

\bibitem[Ke et~al.(2007)Ke, Sukthankar, and Hebert]{ke2007event}
Yan Ke, Rahul Sukthankar, and Martial Hebert.
\newblock Event detection in crowded videos.
\newblock In \emph{Computer Vision, 2007. ICCV 2007. IEEE 11th International
  Conference on}, pages 1--8. IEEE, 2007.

\bibitem[Kim et~al.(2010)Kim, Lee, and Lee]{Kim2010LearningAffinities}
Tae~Hoon Kim, Kyoung~Mu Lee, and Sang~Uk Lee.
\newblock Learning full pairwise affinities for spectral segmentation.
\newblock In \emph{Computer Vision and Pattern Recognition (CVPR), 2010 IEEE
  Conference on}, pages 2101--2108, June 2010.

\bibitem[Kohli et~al.(2009)Kohli, Torr, et~al.]{kohli2009robust}
Pushmeet Kohli, Philip~HS Torr, et~al.
\newblock Robust higher order potentials for enforcing label consistency.
\newblock \emph{International Journal of Computer Vision}, 82\penalty0
  (3):\penalty0 302--324, 2009.

\bibitem[Leibe and Schiele(2004)]{leibe2004scale}
Bastian Leibe and Bernt Schiele.
\newblock Scale-invariant object categorization using a scale-adaptive
  mean-shift search.
\newblock In \emph{Pattern Recognition}, pages 145--153. Springer, 2004.

\bibitem[Martin et~al.(2001)Martin, Fowlkes, Tal, and
  Malik]{martin2001database}
David Martin, Charless Fowlkes, Doron Tal, and Jitendra Malik.
\newblock A database of human segmented natural images and its application to
  evaluating segmentation algorithms and measuring ecological statistics.
\newblock In \emph{Computer Vision, 2001. ICCV 2001. Proceedings. Eighth IEEE
  International Conference on}, volume~2, pages 416--423. IEEE, 2001.

\bibitem[Mayer and Greenspan(2009)]{mayer2009adaptive}
Arnaldo Mayer and Hayit Greenspan.
\newblock An adaptive mean-shift framework for mri brain segmentation.
\newblock \emph{Medical Imaging, IEEE Transactions on}, 28\penalty0
  (8):\penalty0 1238--1250, 2009.

\bibitem[Ozertem et~al.(2008)Ozertem, Erdogmus, and Jenssen]{ozertem2008mean}
Umut Ozertem, Deniz Erdogmus, and Robert Jenssen.
\newblock Mean shift spectral clustering.
\newblock \emph{Pattern Recognition}, 41\penalty0 (6):\penalty0 1924--1938,
  2008.

\bibitem[Paris and Durand(2007)]{paris2007topological}
Sylvain Paris and Fr{\'e}do Durand.
\newblock A topological approach to hierarchical segmentation using mean shift.
\newblock In \emph{Computer Vision and Pattern Recognition, 2007. CVPR'07. IEEE
  Conference on}, pages 1--8. IEEE, 2007.

\bibitem[Shotton et~al.(2013)Shotton, Sharp, Kipman, Fitzgibbon, Finocchio,
  Blake, Cook, and Moore]{shotton2013real}
Jamie Shotton, Toby Sharp, Alex Kipman, Andrew Fitzgibbon, Mark Finocchio,
  Andrew Blake, Mat Cook, and Richard Moore.
\newblock Real-time human pose recognition in parts from single depth images.
\newblock \emph{Communications of the ACM}, 56\penalty0 (1):\penalty0 116--124,
  2013.

\bibitem[Surkala et~al.(2012)Surkala, Mozdren, Fusek, and
  Sojka]{Surkala2012Evolving}
M.~Surkala, K.~Mozdren, R.~Fusek, and E.~Sojka.
\newblock Hierarchical evolving mean-shift.
\newblock In \emph{Image Processing (ICIP), 2012 19th IEEE International
  Conference on}, pages 1593--1596, 2012.

\bibitem[{\v{S}}urkala et~al.(2011){\v{S}}urkala, Mozd{\v{r}}e{\v{n}}, Fusek,
  and Sojka]{vsurkala2011hierarchical}
Milan {\v{S}}urkala, Karel Mozd{\v{r}}e{\v{n}}, Radovan Fusek, and Eduard
  Sojka.
\newblock Hierarchical blurring mean-shift.
\newblock In \emph{Advances Concepts for Intelligent Vision Systems}, pages
  228--238. Springer, 2011.

\bibitem[Unnikrishnan et~al.(2007)Unnikrishnan, Pantofaru, and
  Hebert]{unnikrishnan2007toward}
Ranjith Unnikrishnan, Caroline Pantofaru, and Martial Hebert.
\newblock Toward objective evaluation of image segmentation algorithms.
\newblock \emph{Pattern Analysis and Machine Intelligence, IEEE Transactions
  on}, 29\penalty0 (6):\penalty0 929--944, 2007.

\bibitem[Vedaldi and Soatto(2008)]{vedaldi2008quick}
Andrea Vedaldi and Stefano Soatto.
\newblock Quick shift and kernel methods for mode seeking.
\newblock In \emph{Computer Vision--ECCV 2008}, pages 705--718. Springer, 2008.

\bibitem[Vojir et~al.(2013)Vojir, Noskova, and Matas]{vojir2013robust}
Tomas Vojir, Jana Noskova, and Jiri Matas.
\newblock Robust scale-adaptive mean-shift for tracking.
\newblock In \emph{Image Analysis}, pages 652--663. Springer, 2013.

\bibitem[Wang et~al.(2004)Wang, Thiesson, Xu, and Cohen]{wang2004image}
Jue Wang, Bo~Thiesson, Yingqing Xu, and Michael Cohen.
\newblock Image and video segmentation by anisotropic kernel mean shift.
\newblock In \emph{Computer Vision-ECCV 2004}, pages 238--249. Springer, 2004.

\bibitem[Yang et~al.(2007)Yang, Meer, and Foran]{yang2007multiple}
Lin Yang, Peter Meer, and David~J Foran.
\newblock Multiple class segmentation using a unified framework over mean-shift
  patches.
\newblock In \emph{Computer Vision and Pattern Recognition, 2007. CVPR'07. IEEE
  Conference on}, pages 1--8. IEEE, 2007.

\bibitem[Yuan et~al.(2012)Yuan, Hu, and He]{yuan2012agglomerative}
Xiao-Tong Yuan, Bao-Gang Hu, and Ran He.
\newblock Agglomerative mean-shift clustering.
\newblock \emph{Knowledge and Data Engineering, IEEE Transactions on},
  24\penalty0 (2):\penalty0 209--219, 2012.

\bibitem[Zhang et~al.(2006)Zhang, Kwok, and Tang]{zhang2006accelerated}
Kai Zhang, Jamesk~T Kwok, and Ming Tang.
\newblock Accelerated convergence using dynamic mean shift.
\newblock In \emph{Computer Vision--ECCV 2006}, pages 257--268. Springer, 2006.

\end{thebibliography}

\end{document}